\documentclass[a4paper,conference]{IEEEtran}
\usepackage{color}
\usepackage{tabularx}
\usepackage{amsmath}
\usepackage{subfigure}
\usepackage{caption}
\usepackage{alphabeta}
\usepackage{graphicx}
\usepackage{float}
\usepackage[export]{adjustbox}
\usepackage[table]{xcolor}
\usepackage{multirow}
\usepackage{tabularx}
\usepackage{pdfpages}
\usepackage{amsmath, amsthm, amssymb}
\newsavebox\mysavebox
\usepackage{cite}

\newcommand{\bftab}{\fontseries{b}\selectfont}

\ifCLASSINFOpdf
\else
\fi
\begin{document}
\raggedbottom
%
\title{Rotation Invariant Aerial Image Retrieval with Group Convolutional Metric Learning}


\author{\IEEEauthorblockN{Hyunseung Chung $^{1}$, Woo-Jeoung Nam $^{2}$, and Seong-Whan Lee $^{1, 2}$}
\IEEEauthorblockA{$1.$ Department of Artificial Intelligence, Korea University\\$2.$ Department of Computer and Radio Communications Engineering, Korea University\\
Email: \{hs\_chung, nwj0612, sw.lee\}@korea.ac.kr
}}

%


\maketitle

\begin{abstract}
Remote sensing image retrieval (RSIR) is the process of ranking database images depending on the degree of similarity compared to the query image. As the complexity of RSIR increases due to the diversity in shooting range, angle, and location of remote sensors, there is an increasing demand for methods to address these issues and improve retrieval performance. In this work, we introduce a novel method for retrieving aerial images by merging group convolution with attention mechanism and metric learning, resulting in robustness to rotational variations. For refinement and emphasis on important features, we applied channel attention in each group convolution stage. By utilizing the characteristics of group convolution and channel-wise attention, it is possible to acknowledge the equality among rotated but identically located images. The training procedure has two main steps: (i) training the network with Aerial Image Dataset (AID) for classification, (ii) fine-tuning the network with triplet-loss for retrieval with Google Earth South Korea and NWPU-RESISC45 datasets. Results show that the proposed method performance exceeds other state-of-the-art retrieval methods in both rotated and original environments. Furthermore, we utilize class activation maps (CAM) to visualize the distinct difference of main features between our method and baseline, resulting in better adaptability in rotated environments.
\end{abstract}


%
\IEEEpeerreviewmaketitle

\section{Introduction}
High-resolution remote sensing images are widely available due to advancements in remote sensors and computer technology in the past few years. As the volume of available images increases rapidly, efficient management of large database of images is a growing challenge in the remote sensing community. As a potential solution to effectively manage huge volume of remote sensing images, aerial image retrieval, a method for searching similar images between the query and database, has been studied to replace the time-consuming handcrafted task. The bird view perspective of remote sensing images has large variations of rotation and scale, and is the main obstacle to overcome in order to successfully conduct the task. These variations cause objects in images to be arbitrary oriented \cite{ding2019learning}, and difficult to extract features from or compare similarities to each other during the retrieval process.

In general image retrieval tasks, resolving the rotation issue is not the key factor to success. Most images are captured in human-view (upright) positions, and convolutional neural network (CNN) as a backbone network can accomplish this task. However, remote sensors such as satellites and aircraft shoot photos from diverse directions and positions while orbiting the earth or hovering above target regions. Although two regions share the same altitude and longitude, the rotational and seasonal differences between remote sensing images can cause difficulty in distinguishing the identical regions. Rotation invariance is commonly used for rotational issues in CNNs and widely studied to resolve this problem in various domains such as face detection \cite{maeng2012nighttime, schroff2015facenet, park2013face, bulthoff2003biologically, shi2018real,  kang2014nighttime, xi2002facial}, texture classification \cite{song2017letrist, lee1990translation, pan2017feature, lee1999integrated}, and object detection \cite{ding2019learning, roh2007accurate, cheng2016learning, roh2010view}. These studies show necessity of determining rotational changes exist in fields that must correctly identify equal categories despite rotational variations. Therefore, an image retrieval network robust to rotational differences between query and database images is critical in remote sensing.  

The crucial factor in creating such a retrieval network is applying fine-tuning. The success of CNN classification in ImageNet \cite{krizhevsky2012imagenet} initiated rapid expansion of fine-tuning to other domains such as object recognition and human pose estimation \cite{yang2007reconstruction}. ImageNet fine-tuning methods also expanded to image retrieval, with most tasks using pre-trained ImageNet models for training. However, remote sensing images have a marginal gap compared to datasets such as ImageNet. The images are taken from a bird's eye view, so each image contains various small objects and buildings. Therefore, fine-tuning the network with remote sensing dataset for classification is an important first step. We train the network with AID \cite{xia2017aid}, because of the characteristics of the dataset such as high number of classes, diverse rotational images, and similar attributes to other aerial datasets.  

The second step of the two-step training process is fine-tuning the network for remote sensing image retrieval. we used the recently generated Google Earth South Korea dataset \cite{yun2020coarse} and benchmark aerial dataset NWPU-RESISC45 \cite{cheng2017remote}  to fine-tune the group convolution network \cite{cohen2016group} with channel attention pre-trained in AID. As the main factor of metric learning, triplet loss \cite{hoffer2015deep} is utilized, in which distance of different attributes is kept greater than the distance of same attributes.

In this paper, we propose a novel method for successfully retrieving arbitrary rotated remote sensing images with large variations in rotation, viewpoint, and environment. In order to make the network rotation-invariant, we first apply the group convolution on metric learning. The group convolutions (g-convolution), which shows impressive performance in rotation tasks, share a higher degree of weights by rotating the filters during training. This helps the network generalize to rotated spatial transformations while using approximately the same number of parameters.

Our \textbf{contributions} in this work are as follows:
\begin{itemize}
\item	We propose a retrieval network that is robust to rotational variations. To the best of our knowledge, this is the first attempt to merge both group convolution with channel attention and metric learning to create a single rotation-invariant network for RSIR.
\item	We use a two-step training strategy: (i) training the classification network to contain the high-level features of remote sensing images, (ii) based on high-level features, fine-tuning the network with metric learning approach to learn the discriminative features among identical locations, but different rotated views. 
\item   In both experiments of rotated and general environments, our method shows superior performance compared to other state-of-the-art methods in remote sensing image retrieval. To verify the advantage of our work, we visualize activated feature maps in the intermediate layer to show what Deep Neural Networks (DNNs) mainly focus on, resulting in the network's enhanced ability to search for crucial features in rotated circumstances.
\end{itemize}

\section{Related Work}
In this section, we will first discuss RSIR \cite{tong2019exploiting}, and then review some of the previous works in rotation-invariant CNNs and deep metric learning.

\cite{dieleman2016exploiting, marcos2017rotation} introduces the characteristics of translation equivariant and invariant properties. For the former case, convolution undergoes the same element-wise product with the overlapping window values on the input image. Therefore, any translation of the input image results in the same feature translated by the same amount after convolution. This preserves the amount of translation from before to after the convolution operation. For the latter case, i.e. invariant property, the pooling operation summarizes the nearby outputs of the convolution operation and produces a new final output. This makes small translational changes in the input insignificant to the final outcome. However, this is not the case for other transformations such as scale or rotation. In order to enhance the conventional CNN to be rotation equivariant / invariant, many of the works in recent years focus on this area of research. Spatial transformer networks \cite{jaderberg2015spatial} improves geometric variations by cropping important parts of the input image and installing spatial transformations between each layer. Although this showed positive classification results, the network becomes more prone to overfitting. The work \cite{dieleman2016exploiting} proposes different types of operations to apply transformations on feature maps produced by the CNNs. In TI-pooling \cite{laptev2016ti}, several rotated versions of an image are used to generate transformation invariant features. However, the aforementioned works are limited to consider global transformations of input images. The network is not able to consider subtle transformations of specific regions. This issue is addressed by rotation equivariant vector field networks \cite{marcos2017rotation} which use vector fields and orientation pooling methods to consider the local orientation information of precise regions in images. \cite{cohen2016group} proposes group convolution for intermediate layers of convolution network to exploit rotated filters in full capacity. We carefully address this method in the main text.

The conventional CNN methods for image retrieval contain two steps \cite{zhou2017learning}: Feature extraction and similarity measure. The feature extraction step generates valuable representations from the image, and similarity measure ranks the similarity between the query and database images for retrieval. In the feature extraction step, the features extracted by convolution are summarized by the fully-connected layer as a feature vector with large dimensions. This makes the similarity measure step extremely difficult. Therefore, the deep metric learning method incorporates the two steps together by efficiently training the network for feature extraction through euclidean distance measure between features. The work \cite{sun2014deep} uses deep metric learning for face recognition task. Contrastive loss \cite{chopra2005learning} is utilized to maximize euclidean distance for tuples with similar features and minimize euclidean distance for tuples with different features. An improvement from this loss is the log-ratio loss \cite{kim2019deep}, which goes beyond the binary supervision of contrastive loss and considers continuous relationships for training the network. \cite{hoffer2015deep} discusses triplet-loss which generates three tuples and configures the distance between anchor, positive, and negative tuples depending on the attributes. Detail of this loss is explained in the deep metric learning section.    

Group convolutions \cite{cohen2016group} utilize rotated filters to capture the distinct features in rotated images. Triplet-loss \cite{hoffer2015deep} takes advantage of similarity measures to effectively position the tuples for the retrieval task. Our rotation invariant retrieval network shares both attributes of rotation invariance and retrieval efficiency by consolidating group convolution and triplet loss.     

\begin{figure}[!t]
\begin{center}
\includegraphics[width=1\columnwidth]{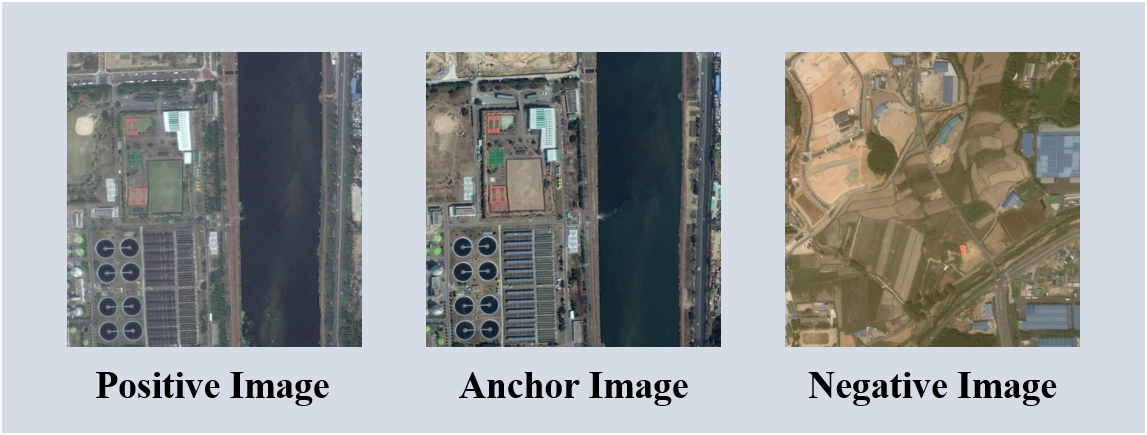}
\end{center}
   \caption{Example of data tuples for remote sensing image retrieval. 
}
\label{fig:1}

\end{figure}

\section{Proposed Method}
In this section, we introduce our method: Group convolutional metric learning network with channel attention for rotation invariant remote sensing image retrieval. The first step is learning the base features by training classification with group convolution neural network (G-CNN). As a similar concept of the fine-tuning process in image classification tasks, the purpose of this stage is for learning the characteristics of aerial images with several rotated filters. Group convolution enables the network to train spatial rotations of the target dataset efficiently by providing the learning mechanisms with various rotated filters. It is possible to retrieve images by utilizing trained features, but not enough to successfully reflect the specific characteristics of the target dataset. In the aerial image domain, a slight difference of viewpoint or time in the same location becomes the main obstacle to overcome for successful retrieval. Therefore, as a solution to the above issue, channel attention and triplet-loss function is adopted with G-CNN to train the network for image retrieval in the second stage. To focus on salient features produced by G-CNN, channel attention module summarizes important feature information and multiplies by input feature maps. Then, by providing a standard of positive and negative images about the anchor image, it is possible to improve the retrieval performance in the aerial dataset.    
\begin{figure*}[!t]
\begin{center}
\includegraphics[width=\linewidth]{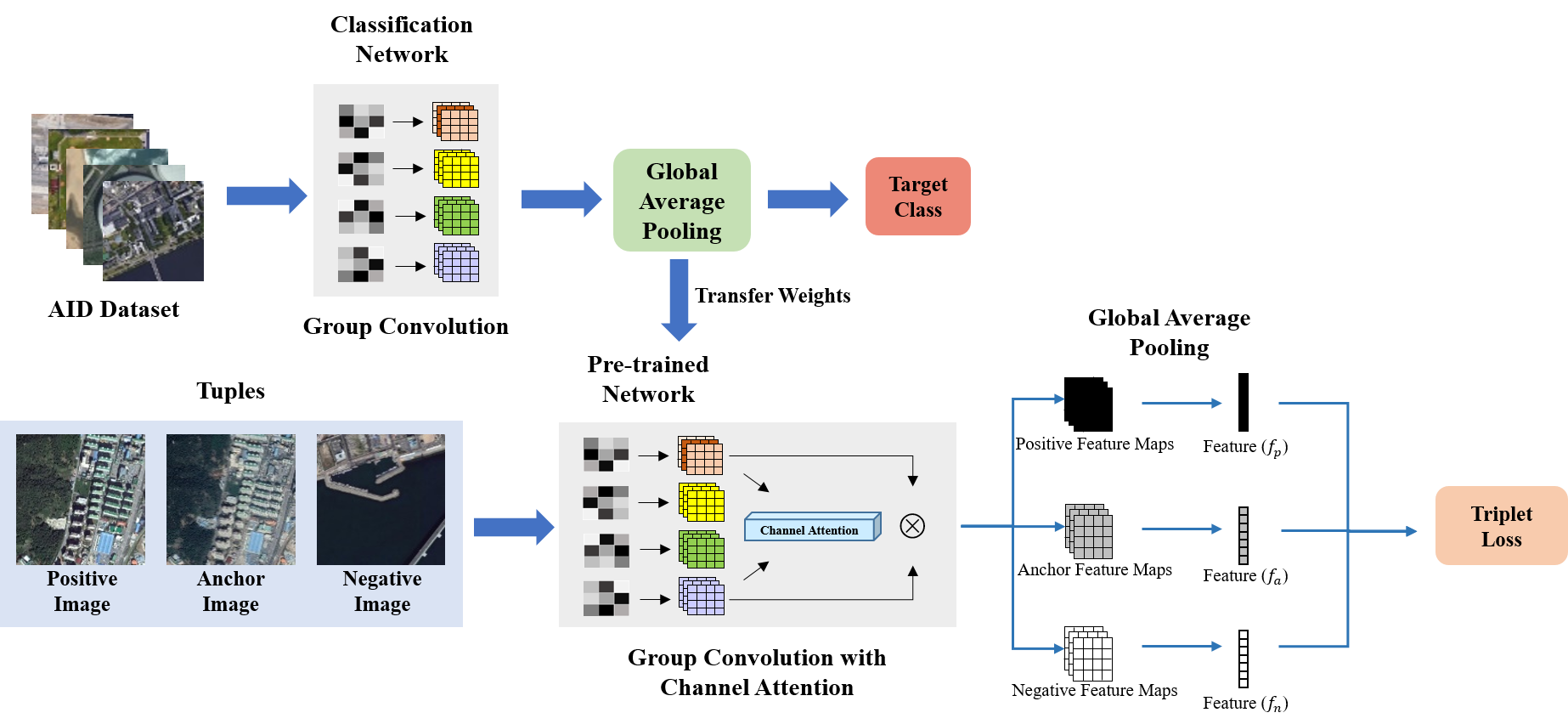}
\end{center}
   \caption{Overall architecture of G-CNN for classification task (\textit{top}) and Group convolutional metric learning network with channel attention for remote sensing image retrieval (\textit{bottom}).     
}
\label{fig:2}

\end{figure*}

\subsection{Group Convolution Neural Network}
In \cite{cohen2016group, veeling2018rotation}, mathematical models of CNNs and G-CNNs are both introduced. A regular convolutional network uses a stack of feature maps $f: Z^2\rightarrow{}R^K$, where K represents the number of channels. The convolution process is defined as follows:

\begin{equation}
\mathcal[f * \psi](x) = \sum_{y \in Z^2} \sum_{k=1}^K f_k(y)\psi_k(x-y)
\end{equation}
Convolution operation: ($\ast$) between the feature maps and filter ($\psi$) is represented in the following way.

G-CNN is an expansion of the CNN by including additional general symmetry groups for convolution. As types of this convolution, $p4$ group includes $90^\circ$ rotations and $p4m$ group additionally contains the reflections of the filters. The feature map outcome of each method is in 4 and 8 times channels compared to the original convolution, respectively. The first layer produces the rotations by the $Z^2\rightarrow{}G$ convolution process:

\begin{equation}
\mathcal[f * \psi](g) = \sum_{y \in Z^2} \sum_{k=1}^K f_k(y)\psi_k(g^{-1}y)
\end{equation}
For $G = p4$ or $G = p4m$, $g = (r,t)$ is a roto-translation or roto-reflection-translation.

In further layers, the convolution process is the same as as the first layer, except the input and output of the convolution function are all in group G:

\begin{equation}
\mathcal[f * \psi](g) = \sum_{h \in G} \sum_{k=1}^K f_k(h)\psi_k(g^{-1}h)
\end{equation}

In summary, equivariant features of G-CNN enables the network to effectively identify rotated aerial images. During group convolution, the filters are transformed instead of the images in order to reduce computational overhead. Keeping the parameters approximately fixed, the input image is convoluted with different rotated filters. Repeatedly, the convoluted product is passed onto the next group equivariant convolutional layer, and convoluted by rotated filters.

As a base model of our work, we utilize ResNet-34 \cite{he2016deep} to integrate the group convolutions. The convolutional layers are replaced by group convolutional layers, and batch normalization layers take into account the additional channel dimensions of the feature maps formed by G-CNNs. This architecture facilitates convolutions for rotated filters within the group symmetry.

\subsection{Channel Attention Module}
Attention mechanism is a powerful method for refining and emphasizing the important feature maps among layers. We utilize a channel attention module to refine the feature maps which contain many spatial transformation information with rotations and reflections after passing Group convolution stage. Motivated by various works in attention such as \cite{hu2018squeeze, woo2018cbam}, we integrate channel attention module with G-CNN. Channel attention considers inter-channel relationships and focuses on critical regions given an input image. Given a input feature map $f_g$ generated by G-CNN, the channel attention module is represented as follows:

\begin{equation}
C_a =\sigma(MLP(P_{avg}(f_g)))
\end{equation}
\begin{equation}
f_r = C_a \otimes f_g
\end{equation}

Here, each notations denote as follow: $\sigma$: Sigmoid function, $MLP$:  multi-layer perceptron, $P_{avg}$: average pooling. The produced channel attention map ($C_a$) undergoes element-wise multiplication ($\otimes$) with input feature map. $f_r$ depicts the final refined feature map for training with triplet loss. 

\subsection{Deep Metric Learning}
As a similar tuple concept in \cite{schroff2015facenet}, three-tuples are constructed for training: anchor, positive, and negative. In Google Earth South Korea dataset, when the anchor image is the target ground truth image, the positive image denotes the same location images, but with time variation of approximately 1 year. The negative image is a completely different region and time. The examples of the three data tuples are shown in Figure \ref{fig:1}.

\cite{hoffer2015deep} introduced triplet-loss function to train the relationships between these tuples. As a standard among tuples, this loss function is able to identify the similarities and differences between the regions. The loss function is defined as follows: 
\begin{equation}
\mathcal{L} = \sum_{i}^{N}[\left|f(x_i^a)-f(x_i^p)\right|_2^2 - \left|f(x_i^a)-f(x_i^n)\right|_2^2 + \alpha]_+
\end{equation}
Here, $x_i^a$, $x_i^p$ and $x_i^n$ denote the anchor image, positive and negative image, respectively. $\alpha$ represents the margin between the positive and negative distances, and $N$ is the total number of samples. The target of this loss is minimizing the relation distance between the anchor and positive (positive distance) while maximizing the distance between the anchor and negative (negative distance). The negative distance is kept greater than the positive distance by a margin $\alpha$. To utilize this loss, it is important to select a variety of triplets for proper training. We use a dense triplet mining method \cite{kim2019deep} to ensure that the greatest variety of triplets are chosen during training.    


\subsection{Overall Network}
As a base step for training the general concepts of aerial images, we train G-CNN to classify AID dataset. Afterward, attentive G-CNN with metric learning is used to fine-tune the network for remote sensing image retrieval. Figure \ref{fig:2} shows the overall training architecture. Google Earth South Korea and NWPU-RESISC45 datasets with original and rotated circumstances are used for the RSIR task. 

\section{Aerial Datasets}
AID is a large-scale aerial image dataset with 30 classes and 10,000 images. AID dataset is collected from Google Earth tools \cite{gorelick2017google}, and the characteristics such as scale, altitude, and quality are similar to Google Earth South Korea dataset. The images are $600 \times 600$ pixels, and the dataset is for aerial scene classification. Instead of the conventional method of using pre-trained ImageNet \cite{deng2009imagenet} for retrieval, we pre-trained the network with AID for retrieval to alleviate the large domain gap. 

The Google Earth South Korea dataset is a multi-temporal dataset, and consist of images of South Korea excluding the islands. The purpose of this dataset is for RSIR tasks, and the collected images are in the time span between 2016 and 2019. There are a total of 40,000 images, and the satellite images are $1080 \times 1080$ composed of buildings, grasslands, rivers, and many other scenes. The images taken each year were of the same areas. 

The NWPU-RESISC45 dataset is public accessible large-scale aerial image dataset with 45 classes and 31,500 images in $256 \times 256$ pixels. It holds variations such as translation, spatial resolution, viewpoint, object pose, illumination, background, and occlusion. Although the purpose of this dataset is for scene classification, this dataset is also suitable for RSIR due to high within-class diversity and between-class similarity.

Both datasets are used for evaluation and comparison between different RSIR methods.

\begin{figure}[t]
\centering     
\subfigure{\includegraphics[width=17mm]{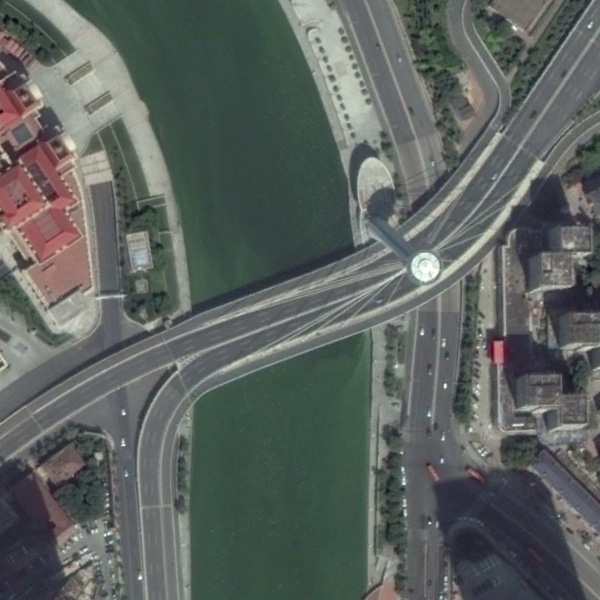}}
\subfigure{\includegraphics[width=17mm]{images/google/beach.jpg}}
\subfigure{\includegraphics[width=17mm]{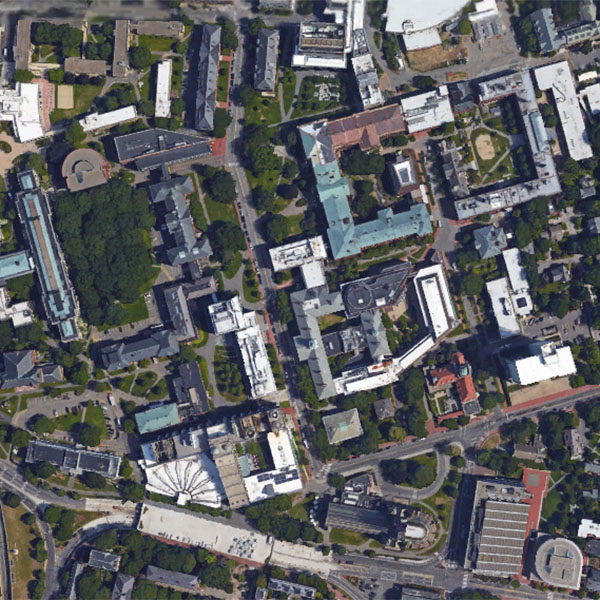}}
\subfigure{\includegraphics[width=17mm]{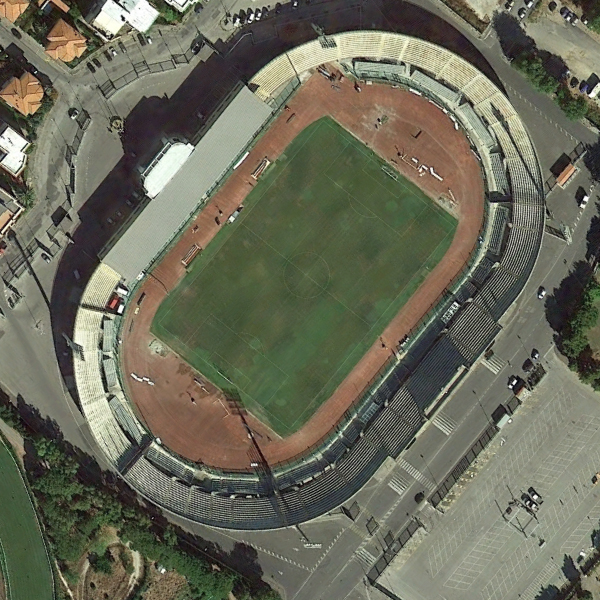}}
\subfigure{\includegraphics[width=17mm]{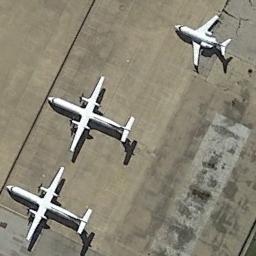}}
\subfigure{\includegraphics[width=17mm]{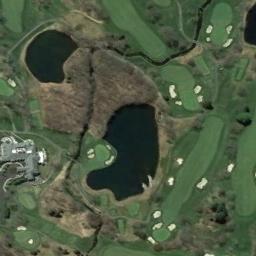}}
\subfigure{\includegraphics[width=17mm]{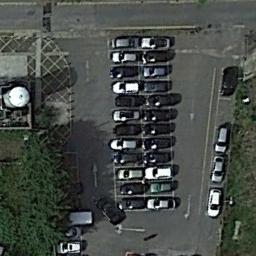}}
\subfigure{\includegraphics[width=17mm]{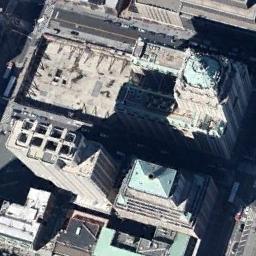}}
\subfigure{\includegraphics[width=17mm]{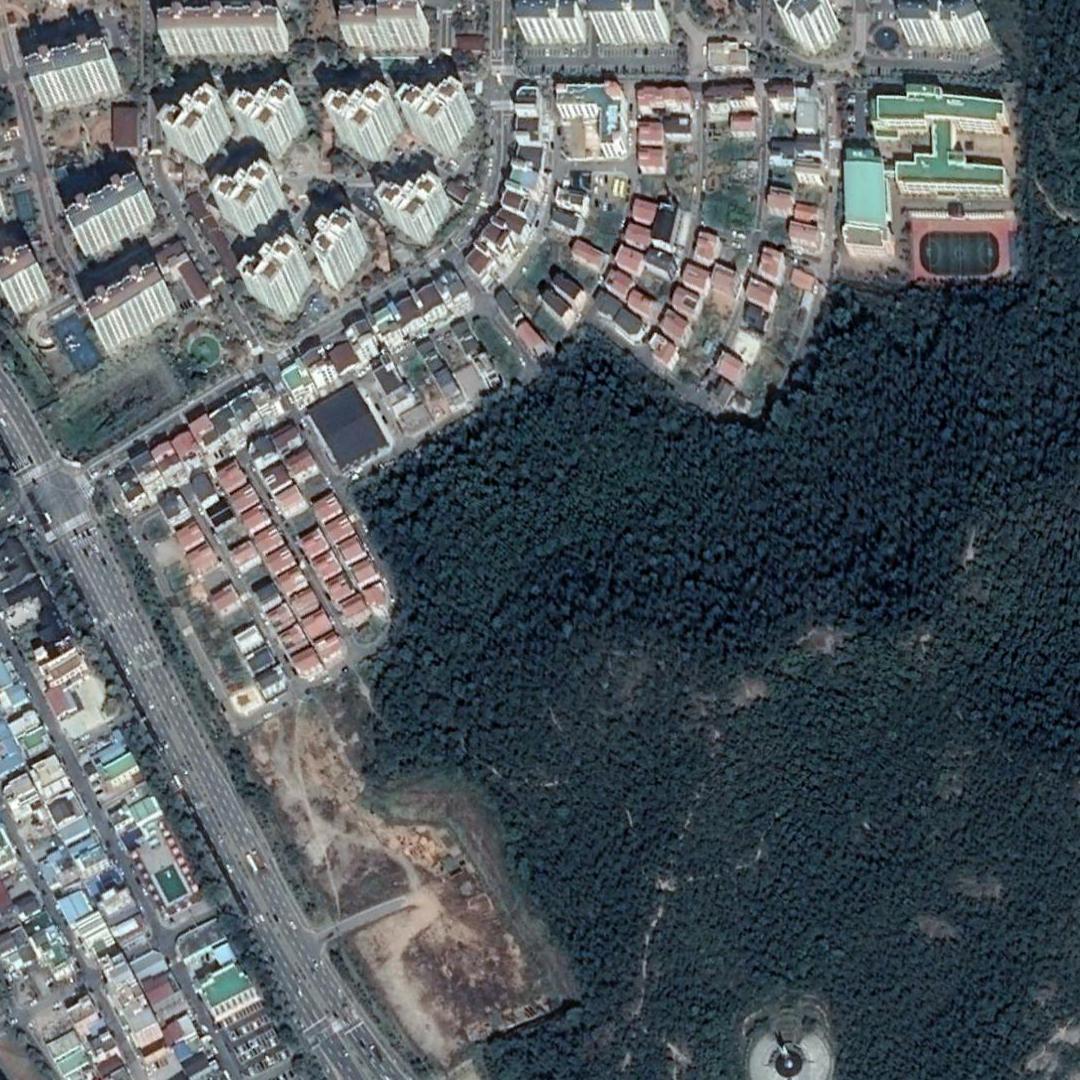}}
\subfigure{\includegraphics[width=17mm]{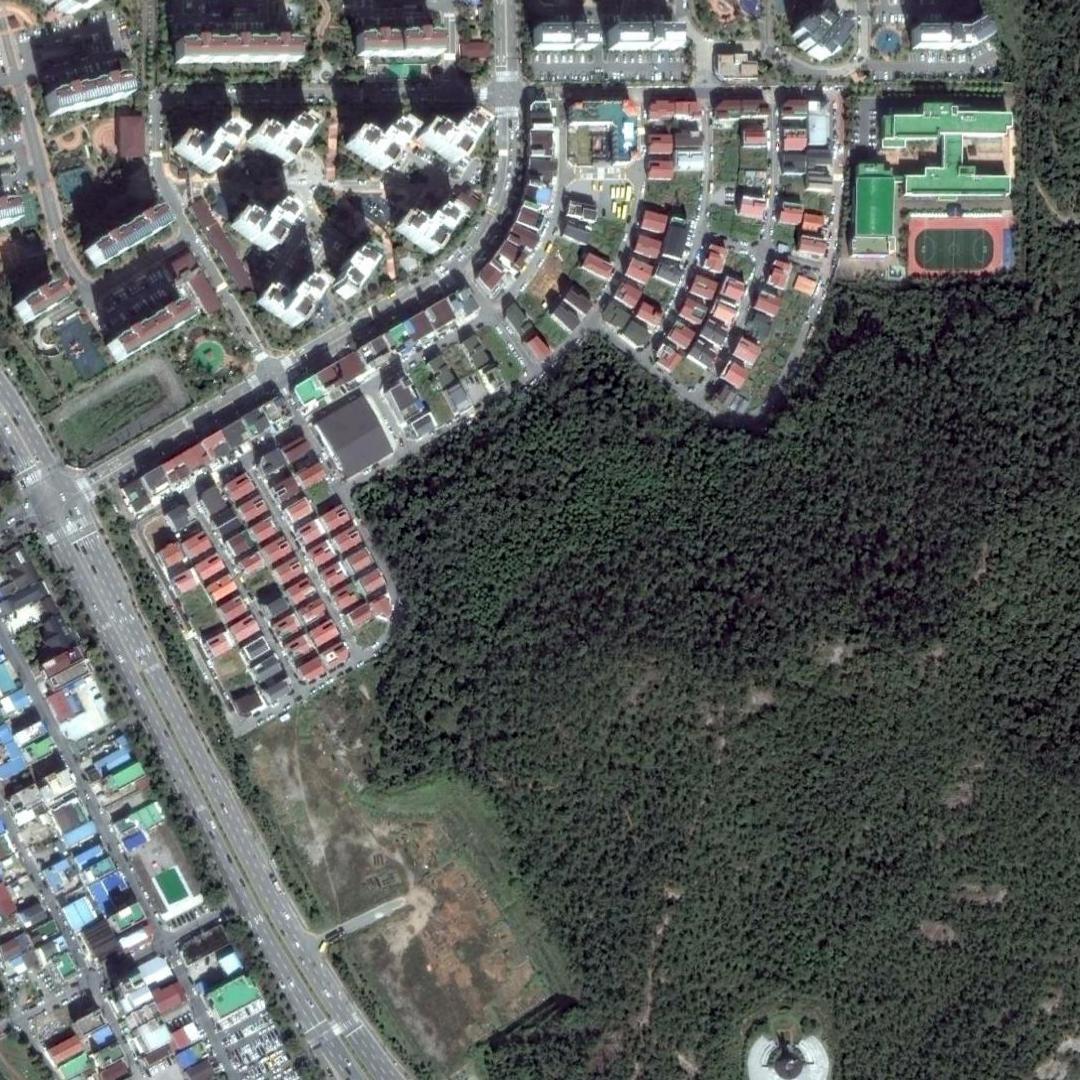}}
\subfigure{\includegraphics[width=17mm]{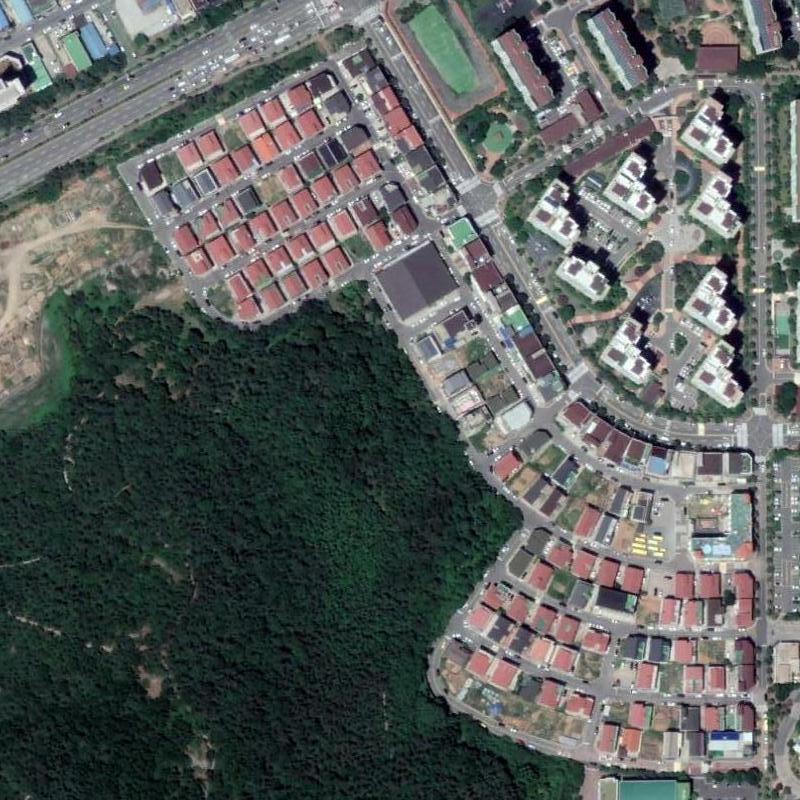}}
\subfigure{\includegraphics[width=17mm]{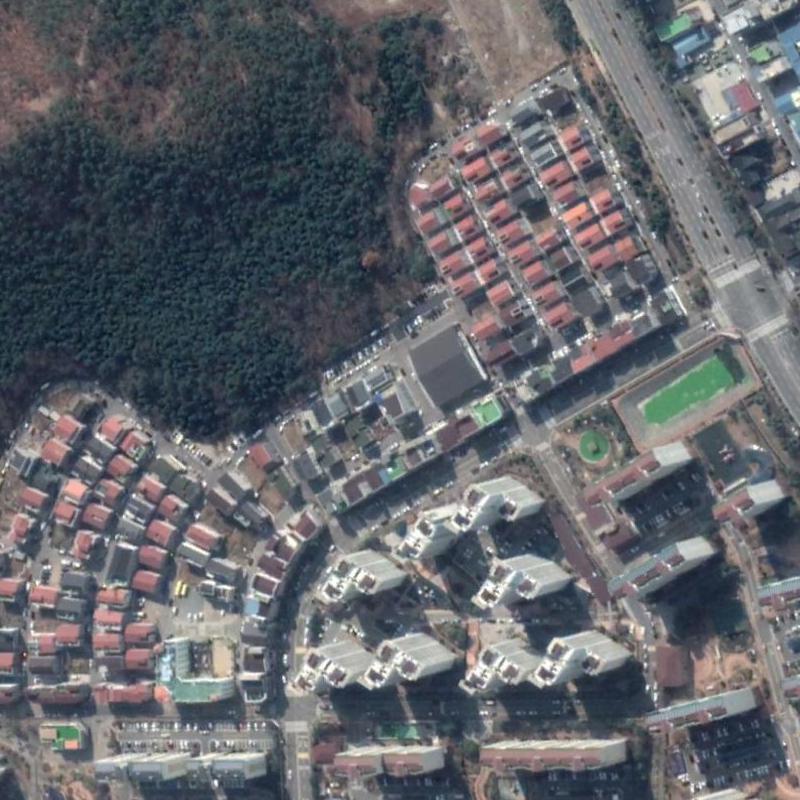}}
\caption{Example image from AID (\textit{top}), NWPU-RESISC45 (\textit{middle}), and Google Earth South Korea dataset with different viewpoint (\textit{bottom}).} 
\label{fig:3}
\end{figure}

To verify our work, we rotated the test set in Google Earth South Korea and NWPU-RESISC45 datasets. For Google Earth South Korea, the 2018 and 2019 images are randomly rotated by 0$^\circ$, 90$^\circ$, 180$^\circ$, or 270$^\circ$ to set the positive image a randomly rotated version of the original image during retrieval testing. The 2016 and 2017 images are utilized as the training dataset. However, for a fair comparison, other methods in the experiment are trained for rotation by data augmentation, and performance in this rotated dataset was evaluated. Examples of images in each dataset are shown in Figure \ref{fig:3}. For NWPU-RESISC45 dataset, random rotation is applied to all classes.  
 
\section{Experiments}
\subsection{Implementation Details}
As a base model, we utilize ResNet-34 architecture, replacing the convolution layers in each of the residual blocks with group convolution layers, and adjusting the batch normalization layers from 2d to 3d structure to take into account the added channels from group convolution. Also, depending on the type of group convolution ($p4$ or $p4m$), the number of parameters drastically increase for each layer because the number of feature maps multiply (4 or 8). To prevent the changes in the number of parameters, the number of filters are halved for each $p4-conv$ layer and divided by $\sqrt{8}$ for each $p4m-conv$ layer \cite{cohen2016group}. This method keeps the number of parameters similar to the original ResNet-34 without group convolution layers.

For training the network with classification, the AID is split into train and validation sets in a ratio of $7:3$. G-CNN with backbone of ResNet-34 is used with both p4 and p4m group convolutions, respectively. Minimized loss function is cross entropy loss, and the model is optimized by stochastic gradient descent. Learning rate is set to $0.01$ with a momentum of $0.9$. 

Afterwards, the network is fine-tuned for RSIR with G-CNN channel attention and triplet-loss. Half of the Google Earth South Korea dataset (2016, 2017) is used for training and the remaining half (2018, 2019) is used for retrieval. The training dataset is split into train ($90\%$) and validation ($10\%$) sets. For the retrieval dataset, 2018 is used as database and approximately 400 images of 2019 are used as query images. A key factor in the selection process of query images is determining images with noticeable and distinct attributes compared to each other. We exclude images consisting of only one type of scene (e.g., grasslands, barren lands, seas). Implementation details are similar to previous training for classification in AID, except the learning rate is set to $0.001$.

For NWPU-RESISC45 dataset, the train and validation sets are also split into $90\%$ and $10\%$. Our model is trained for $30$ epochs and the learning rate is set to $10^{-4}$. 

\subsection{Performance Evaluation}
The retrieval performance for Google Earth South Korea dataset is evaluated by calculating the correctly retrieved queries ($Recall@n$). If any of the top \textit{n} retrieved database images overlap $50\%$ in contents with the query image, the query image is considered correctly retrieved. The correctly retrieved queries are represented in percentages, and the performance is evaluated for \textit{n} = 1, 5, 10, and 100. The retrieval performance for NWPU-RESISC45 dataset is also evaluated by calculating the top retrieved queries ($Recall@n$), and the performance is evalulated for \textit{n} = 1, 2, 4, and 8.       

\subsection{Experimental Results} 
The RSIR performance of group convolutional metric learning network is compared to other state-of-the-art methods in Google Earth South Korea dataset (Table \ref{table:1}) and NWPU-RESISC45 dataset (Table \ref{table:3}). The methods we use for this comparison are R-MAC descriptor \cite{tolias2015particular}, NetVLAD \cite{arandjelovic2016netvlad}, LDCNN \cite{zhou2017learning},  contrastive loss, triplet loss, G-CNN with contrastive loss (p4m), attentive G-CNN with contrastive loss (p4m), and G-CNN with triplet loss (p4, p4m). The R-MAC descriptor uses a three-stream siamese network and produces global representation of images by aggregating many activation regions. NetVLAD is largely used for place recognition and is still a baseline network for such tasks. LDCNN is actively used to learn low dimensional features in remote sensing images. In order to make the comparison fair, all methods are trained and tested in the same conditions. The methods are pre-trained for classification with AID , and then fine-tuned for RSIR. Also, the backbone architecture for all methods is ResNet-34, and the extracted feature dimensions are 512. The $\ast$ represents channel attention added to each layer of G-CNN. The proposed method confirms the best performance in both datasets.

\begin{table}[t]
   \caption{Retrieval results of state-of-the-art methods on the Google Earth South Korea dataset ($Recall@n$).}
   \label{table:1}
   \centering 
   \begin{tabular}{ p{3.4cm}   p{1.6cm} p{1.6cm} p{1.6cm} p{1.6cm}}
   \rowcolor[HTML]{EFEFEF} 
    \hline
\rowcolor[HTML]{EFEFEF} 
\cellcolor[HTML]{EFEFEF}                         & \multicolumn{4}{c}{\cellcolor[HTML]{EFEFEF}$Recall@n (\%)$} \\ \cline{2-5} 
\rowcolor[HTML]{EFEFEF} 
\multicolumn{1}{c}{\multirow{-2}{*}{\cellcolor[HTML]{EFEFEF}\textbf{Methods}}} & \multicolumn{1}{c}{$n=1$}         &\multicolumn{1}{c}{$n=5$}        & \multicolumn{1}{c}{$n=10$}        & \multicolumn{1}{c}{$n=100$}        \\
\multicolumn{1}{c}{R-MAC descriptor}                       &\multicolumn{1}{c}{14.7}     &\multicolumn{1}{c}{23.9}     &\multicolumn{1}{c}{31.8}       &\multicolumn{1}{c}{46.7}     \\ \hline
\multicolumn{1}{c}{NetVLAD}                   &\multicolumn{1}{c}{15.2}      &\multicolumn{1}{c}{26.8}     &\multicolumn{1}{c}{33.3}       &\multicolumn{1}{c}{55.4}     \\ \hline
\multicolumn{1}{c}{Contrastive loss}                                  &\multicolumn{1}{c}{16.8}     &\multicolumn{1}{c}{27.9}     &\multicolumn{1}{c}{34.5}       &\multicolumn{1}{c}{51.5}     \\ \hline
\multicolumn{1}{c}{Triplet loss}                              &\multicolumn{1}{c}{18.8}     &\multicolumn{1}{c}{29.3}     &\multicolumn{1}{c}{33.7}       &\multicolumn{1}{c}{51.7}     \\ \hline
\multicolumn{1}{c}{LDCNN}                              &\multicolumn{1}{c}{19.2}     &\multicolumn{1}{c}{29.5}     &\multicolumn{1}{c}{34.2}       &\multicolumn{1}{c}{51.5}     \\ \hline
\multicolumn{1}{c}{G-CNN (p4m) + Cont. loss}                              &\multicolumn{1}{c}{22.1}     &\multicolumn{1}{c}{31.4}     &\multicolumn{1}{c}{36.5}       &\multicolumn{1}{c}{54.8}     \\ \hline
\multicolumn{1}{c}{G-CNN (p4m) + Cont. $\ast$}                              &\multicolumn{1}{c}{22.8}     &\multicolumn{1}{c}{31.6}     &\multicolumn{1}{c}{36.9}       &\multicolumn{1}{c}{55.6}     \\ \hline
\multicolumn{1}{c}{G-CNN (p4) + Triplet loss}                              &\multicolumn{1}{c}{27.0}     &\multicolumn{1}{c}{36.0}     &\multicolumn{1}{c}{41.6}       &\multicolumn{1}{c}{65.0}     \\ \hline
\multicolumn{1}{c}{G-CNN (p4) + Triplet. $\ast$}                              &\multicolumn{1}{c}{27.6}     &\multicolumn{1}{c}{36.4}     &\multicolumn{1}{c}{42.2}       &\multicolumn{1}{c}{64.8}     \\ \hline
\multicolumn{1}{c}{G-CNN (p4m) + Triplet loss}                              &\multicolumn{1}{c}{28.1}     &\multicolumn{1}{c}{36.8}     &\multicolumn{1}{c}{42.9}       &\multicolumn{1}{c}{67.3}     \\ \hline
\multicolumn{1}{c}{\bftab G-CNN (p4m) + Triplet. $\boldsymbol\ast$}                              &\multicolumn{1}{c}{\bftab 28.3}     &\multicolumn{1}{c}{\bftab 36.9}     &\multicolumn{1}{c}{\bftab 43.2}      &\multicolumn{1}{c}{67.3}     \\ \hline
   \end{tabular}
\end{table}

\begin{table}[t]
   \caption{Retrieval results of state-of-the-art methods on the rotated Google Earth South Korea dataset ($Recall@n$).}
   \label{table:2}
   \centering 
   \begin{tabular}{p{3.4cm}   p{1.6cm} p{1.6cm} p{1.6cm} p{1.6cm}}
   \rowcolor[HTML]{EFEFEF} 
    \hline
\rowcolor[HTML]{EFEFEF} 
\cellcolor[HTML]{EFEFEF}                         & \multicolumn{4}{c}{\cellcolor[HTML]{EFEFEF}$Recall@n (\%)$} \\ \cline{2-5} 
\rowcolor[HTML]{EFEFEF} 
\multicolumn{1}{c}{\multirow{-2}{*}{\cellcolor[HTML]{EFEFEF}\textbf{Methods}}} & \multicolumn{1}{c}{$n=1$}         &\multicolumn{1}{c}{$n=5$}        & \multicolumn{1}{c}{$n=10$}        & \multicolumn{1}{c}{$n=100$}        \\
\multicolumn{1}{c}{R-MAC descriptor $\dagger$}                       &\multicolumn{1}{c}{6.5}     &\multicolumn{1}{c}{14.5}     &\multicolumn{1}{c}{24.8}       &\multicolumn{1}{c}{64.0}     \\ \hline
\multicolumn{1}{c}{NetVLAD $\dagger$}                   &\multicolumn{1}{c}{7.4}      &\multicolumn{1}{c}{17.4}     &\multicolumn{1}{c}{25.1}       &\multicolumn{1}{c}{68.2}     \\ \hline
\multicolumn{1}{c}{Contrastive loss $\dagger$}                                  &\multicolumn{1}{c}{8.1}     &\multicolumn{1}{c}{16.8}     &\multicolumn{1}{c}{24.5}       &\multicolumn{1}{c}{65.0}     \\ \hline
\multicolumn{1}{c}{Triplet loss $\dagger$}                              &\multicolumn{1}{c}{6.9}     &\multicolumn{1}{c}{18.0}     &\multicolumn{1}{c}{24.7}       &\multicolumn{1}{c}{65.6}     \\ \hline
\multicolumn{1}{c}{LDCNN $\dagger$}                              &\multicolumn{1}{c}{8.9}     &\multicolumn{1}{c}{18.4}     &\multicolumn{1}{c}{24.7}       &\multicolumn{1}{c}{66.6}     \\ \hline
\multicolumn{1}{c}{G-CNN (p4m) + Cont. loss}                              &\multicolumn{1}{c}{17.8}     &\multicolumn{1}{c}{32.4}     &\multicolumn{1}{c}{38.9}       &\multicolumn{1}{c}{72.0}     \\ \hline
\multicolumn{1}{c}{G-CNN (p4m) + Cont. $\ast$}                              &\multicolumn{1}{c}{18.5}     &\multicolumn{1}{c}{32.8}     &\multicolumn{1}{c}{39.8}       &\multicolumn{1}{c}{72.0}     \\ \hline
\multicolumn{1}{c}{G-CNN (p4) + Triplet loss}                              &\multicolumn{1}{c}{20.1}     &\multicolumn{1}{c}{36.0}     &\multicolumn{1}{c}{43.2}       &\multicolumn{1}{c}{76.9}     \\ \hline
\multicolumn{1}{c}{G-CNN (p4) + Triplet. $\ast$}                              &\multicolumn{1}{c}{21.2}     &\multicolumn{1}{c}{36.4}     &\multicolumn{1}{c}{43.9}       &\multicolumn{1}{c}{77.2}     \\ \hline
\multicolumn{1}{c}{G-CNN (p4m) + Triplet loss}                              &\multicolumn{1}{c}{23.4}     &\multicolumn{1}{c}{44.0}     &\multicolumn{1}{c}{51.7}       &\multicolumn{1}{c}{84.6}     \\ \hline
\multicolumn{1}{c}{\bftab G-CNN (p4m) + Triplet. $\boldsymbol\ast$}                              &\multicolumn{1}{c}{\bftab 24.5}     &\multicolumn{1}{c}{\bftab 46.9}     &\multicolumn{1}{c}{\bftab 52.8}      &\multicolumn{1}{c}{\bftab 86.3}     \\ \hline
   \end{tabular}
\end{table}

The RSIR performance is evaluated in rotated Google Earth South Korea and NWPU-RESISC45 dataset. Experiments are done to evaluate the retrieval performance of various methods in datasets with rotational variations as shown in Table \ref{table:2}, \ref{table:4}. In order to validate the experiments, the methods are trained rotation through data augmentation, and the $\dagger$ shows such training has been done. In both datasets, the images are given random rotations in [0$^\circ$, 90$^\circ$, 180$^\circ$, 270$^\circ$]. The experimental results show that our attentive group convolution triplet loss network with p4m group layers has the best retrieval performance compared to other state-of-the-art methods.

\begin{figure*} [t]
\centering
  \includegraphics[width=1.8\columnwidth]{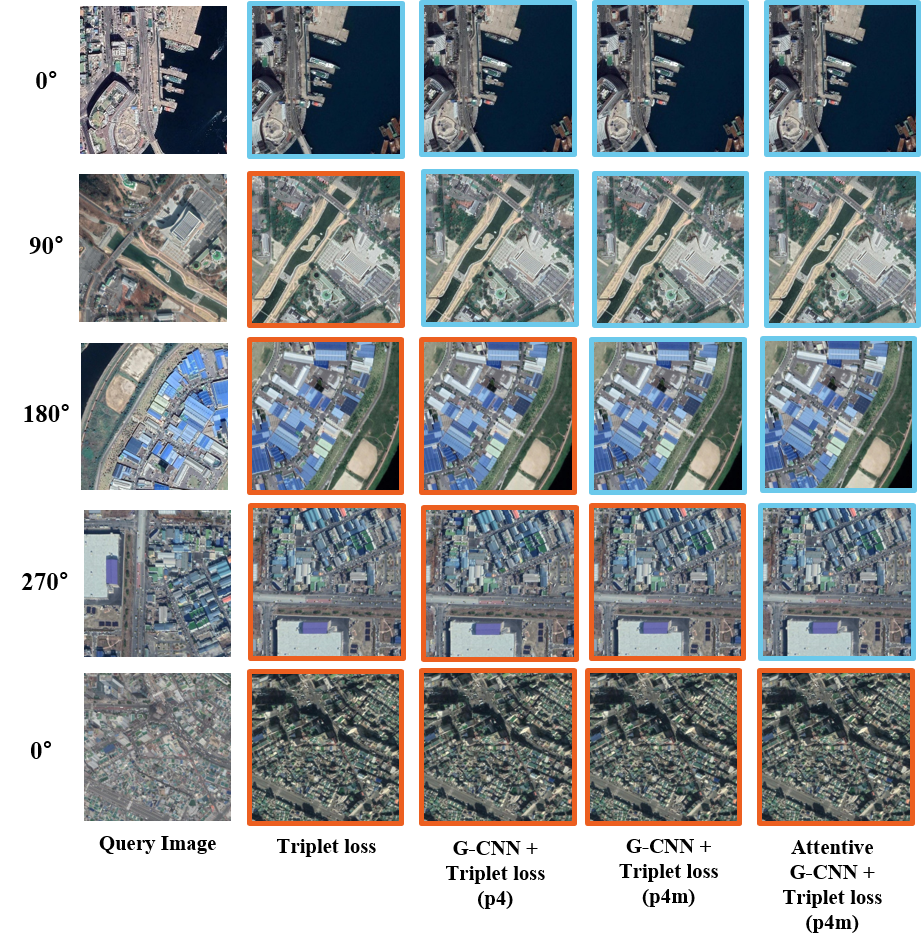}
  \caption{Examples of retrieval results in the rotated Google Earth South Korea dataset. Each column represents different methods and each row represents rotation angle differences in the clockwise direction between query and database images. Blue borders represent correctly retrieved images, and orange borders represent images unsuccessfully retrieved by the methods.}
\label{fig:5}
\end{figure*}

Figure \ref{fig:5} shows examples of retrieval results for variational methods of triplet loss in the rotated Google Earth South Korea dataset. the first column is the query image used for RSIR. The second column is retrieval results with triplet loss. The third column is G-CNN with triplet loss using p4 group convolution layers. The fourth column is G-CNN with triplet loss using p4m, and the last column is attentive G-CNN with triplet loss using p4m group convolutional layers. Attentive G-CNN with triplet loss using p4m group convolution layers maintains reliable retrieval results compared to other methods. 

Proposed method shows similar performance in NWPU-RESISC45 dataset compared to other methods. A key difference of NWPU-RESISC45 over Google Earth South Korea dataset is the main objective of retrieval. In NWPU-RESISC45 dataset, the aim is to return images in the same class as the query image. In Google Earth South Korea dataset, the goal is to return a image in the same location as the query image. This is the reason retrieval performance is relatively higher in original and rotated NWPU-RESISC45 dataset compared to original and rotated Google Earth South Korea dataset.  

\begin{table}[t]
   \caption{Retrieval results of state-of-the-art methods on the NWPU-RESISC45 dataset ($Recall@n$).}
   \label{table:3}
   \centering 
   \begin{tabular}{ p{3.6cm}   p{1.6cm} p{1.6cm} p{1.6cm} p{1.6cm}}
   \rowcolor[HTML]{EFEFEF} 
    \hline
\rowcolor[HTML]{EFEFEF} 
\cellcolor[HTML]{EFEFEF}                         & \multicolumn{4}{c}{\cellcolor[HTML]{EFEFEF}$Recall@n (\%)$} \\ \cline{2-5} 
\rowcolor[HTML]{EFEFEF} 
\multicolumn{1}{c}{\multirow{-2}{*}{\cellcolor[HTML]{EFEFEF}\textbf{Methods}}} & \multicolumn{1}{c}{$n=1$}         &\multicolumn{1}{c}{$n=2$}        & \multicolumn{1}{c}{$n=4$}        & \multicolumn{1}{c}{$n=8$}        \\
\multicolumn{1}{c}{R-MAC descriptor}                       &\multicolumn{1}{c}{25.4}     &\multicolumn{1}{c}{43.3}     &\multicolumn{1}{c}{65.8}       &\multicolumn{1}{c}{85.7}     \\ \hline
\multicolumn{1}{c}{Contrastive loss}                   &\multicolumn{1}{c}{26.6}      &\multicolumn{1}{c}{44.8}     &\multicolumn{1}{c}{67.0}       &\multicolumn{1}{c}{85.7}     \\ \hline
\multicolumn{1}{c}{NetVLAD}                                  &\multicolumn{1}{c}{28.8}     &\multicolumn{1}{c}{46.7}     &\multicolumn{1}{c}{68.3}       &\multicolumn{1}{c}{85.7}     \\ \hline
\multicolumn{1}{c}{Triplet loss}                              &\multicolumn{1}{c}{29.6}     &\multicolumn{1}{c}{48.3}     &\multicolumn{1}{c}{70.8}       &\multicolumn{1}{c}{87.7}     \\ \hline
\multicolumn{1}{c}{LDCNN}                              &\multicolumn{1}{c}{36.6}     &\multicolumn{1}{c}{54.6}     &\multicolumn{1}{c}{72.8}       &\multicolumn{1}{c}{88.2}     \\ \hline
\multicolumn{1}{c}{G-CNN (p4m) + Cont. loss}                              &\multicolumn{1}{c}{39.9}     &\multicolumn{1}{c}{58.1}     &\multicolumn{1}{c}{75.9}       &\multicolumn{1}{c}{88.8}     \\ \hline
\multicolumn{1}{c}{G-CNN (p4m) + Cont. $\ast$}                              &\multicolumn{1}{c}{40.6}     &\multicolumn{1}{c}{58.5}     &\multicolumn{1}{c}{75.3}       &\multicolumn{1}{c}{88.7}     \\ \hline
\multicolumn{1}{c}{G-CNN (p4) + Triplet loss}                              &\multicolumn{1}{c}{42.5}     &\multicolumn{1}{c}{60.0}     &\multicolumn{1}{c}{76.5}       &\multicolumn{1}{c}{88.3}     \\ \hline
\multicolumn{1}{c}{G-CNN (p4m) + Triplet loss}                              &\multicolumn{1}{c}{43.0}     &\multicolumn{1}{c}{60.5}     &\multicolumn{1}{c}{76.5}       &\multicolumn{1}{c}{89.1}     \\ \hline
\multicolumn{1}{c}{G-CNN (p4) + Triplet. $\ast$}                              &\multicolumn{1}{c}{43.1}     &\multicolumn{1}{c}{\bftab 62.7}     &\multicolumn{1}{c}{78.1}       &\multicolumn{1}{c}{89.1}     \\ \hline
\multicolumn{1}{c}{\bftab G-CNN (p4m) + Triplet. $\boldsymbol\ast$}                              &\multicolumn{1}{c}{\bftab 44.5}     &\multicolumn{1}{c}{62.5}     &\multicolumn{1}{c}{\bftab 78.5}      &\multicolumn{1}{c}{\bftab 89.9}     \\ \hline
   \end{tabular}
\end{table}

\begin{table}[t]
   \caption{Retrieval results of state-of-the-art methods on the rotated NWPU-RESISC45 dataset ($Recall@n$).}
   \label{table:4}
   \centering 
   \begin{tabular}{p{3.6cm}   p{1.6cm} p{1.6cm} p{1.6cm} p{1.6cm}}
   \rowcolor[HTML]{EFEFEF} 
    \hline
\rowcolor[HTML]{EFEFEF} 
\cellcolor[HTML]{EFEFEF}                         & \multicolumn{4}{c}{\cellcolor[HTML]{EFEFEF}$Recall@n (\%)$} \\ \cline{2-5} 
\rowcolor[HTML]{EFEFEF} 
\multicolumn{1}{c}{\multirow{-2}{*}{\cellcolor[HTML]{EFEFEF}\textbf{Methods}}} & \multicolumn{1}{c}{$n=1$}         &\multicolumn{1}{c}{$n=2$}        & \multicolumn{1}{c}{$n=4$}        & \multicolumn{1}{c}{$n=8$}        \\
\multicolumn{1}{c}{R-MAC descriptor $\dagger$}                       &\multicolumn{1}{c}{23.1}     &\multicolumn{1}{c}{41.4}     &\multicolumn{1}{c}{62.4}       &\multicolumn{1}{c}{83.3}     \\ \hline
\multicolumn{1}{c}{Contrastive loss $\dagger$}                   &\multicolumn{1}{c}{26.6}      &\multicolumn{1}{c}{46.1}     &\multicolumn{1}{c}{66.7}       &\multicolumn{1}{c}{86.4}     \\ \hline
\multicolumn{1}{c}{NetVLAD $\dagger$}                                  &\multicolumn{1}{c}{27.1}     &\multicolumn{1}{c}{45.3}     &\multicolumn{1}{c}{66.7}       &\multicolumn{1}{c}{85.5}     \\ \hline
\multicolumn{1}{c}{Triplet loss $\dagger$}                              &\multicolumn{1}{c}{35.3}     &\multicolumn{1}{c}{53.1}     &\multicolumn{1}{c}{73.2}       &\multicolumn{1}{c}{88.7}     \\ \hline
\multicolumn{1}{c}{LDCNN $\dagger$}                              &\multicolumn{1}{c}{36.0}     &\multicolumn{1}{c}{54.5}     &\multicolumn{1}{c}{74.0}       &\multicolumn{1}{c}{88.8}     \\ \hline
\multicolumn{1}{c}{G-CNN (p4m) + Cont. loss}                              &\multicolumn{1}{c}{39.9}     &\multicolumn{1}{c}{56.6}     &\multicolumn{1}{c}{74.9}       &\multicolumn{1}{c}{88.3}     \\ \hline
\multicolumn{1}{c}{G-CNN (p4m) + Cont. $\ast$}                              &\multicolumn{1}{c}{41.1}     &\multicolumn{1}{c}{59.8}     &\multicolumn{1}{c}{76.9}       &\multicolumn{1}{c}{89.3}     \\ \hline
\multicolumn{1}{c}{G-CNN (p4) + Triplet loss}                              &\multicolumn{1}{c}{43.1}     &\multicolumn{1}{c}{60.8}     &\multicolumn{1}{c}{77.6}       &\multicolumn{1}{c}{89.5}     \\ \hline
\multicolumn{1}{c}{G-CNN (p4) + Triplet. $\ast$}                              &\multicolumn{1}{c}{43.4}     &\multicolumn{1}{c}{61.3}     &\multicolumn{1}{c}{77.7}       &\multicolumn{1}{c}{89.2}     \\ \hline
\multicolumn{1}{c}{G-CNN (p4m) + Triplet loss}                              &\multicolumn{1}{c}{44.8}     &\multicolumn{1}{c}{62.1}     &\multicolumn{1}{c}{77.8}       &\multicolumn{1}{c}{89.5}     \\ \hline
\multicolumn{1}{c}{\bftab G-CNN (p4m) + Triplet. \bftab $\boldsymbol\ast$}                              &\multicolumn{1}{c}{\bftab 45.7}     &\multicolumn{1}{c}{\bftab 64.3}     &\multicolumn{1}{c}{\bftab 80.2}      &\multicolumn{1}{c}{89.5}     \\ \hline
   \end{tabular}
\end{table}

\begin{figure}[t]
\begin{center}
\includegraphics[width=1\columnwidth]{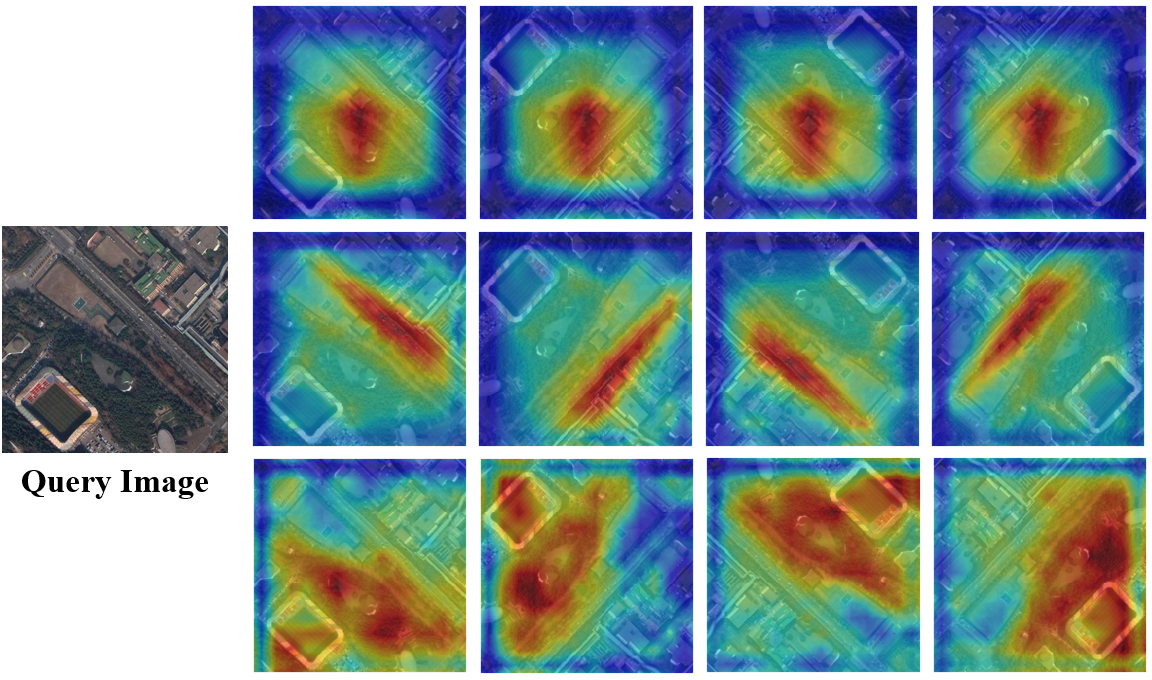}
\end{center}
   \caption{Each columns represent CAM of query image rotated by 0$^\circ$, 90$^\circ$, 180$^\circ$, 270$^\circ$ from left to right. (i) CAM with CNN and triplet-loss (\textit{top}), (ii) CAM with group convolution triplet loss network (p4m) (\textit{middle}), and (iii) CAM with attentive group convolution triplet loss network (p4m) (\textit{bottom})} 
\label{fig:4}
\end{figure}

\subsection{Class Activation Analysis}
We provide visualizations of class activation regions to help better understand the effectiveness of attentive group convolution triplet loss network. The output of the last group convolution (or convolution) layer is shown in Figure \ref{fig:4}, and red represents highly activated regions while blue represent low activation regions. In the original paper that introduced CAM \cite{zhou2016learning}, the activation mappings expressed the discriminative regions used by the network to identify a particular class. However, because the task at hand is retrieval, mappings indicate regions utilized by the network to retrieve a particular database image. Rotations of the query image show clear adjustments in the discriminative regions for group convolution triplet loss network (p4m), and enhanced regions for attentive group convolution triplet loss network (p4m). On the other hand, activation regions stay relatively similar for CNN with triplet loss. This shows the greater expressive capacity in the rotational circumstances of attentive group convolution triplet loss network.        

\section{Discussion}
We apply group convolution layers to train the network for classification with AID. Then, we utilize the datasets to fine-tune the network for RSIR. The performance is evaluated for both Google Earth South Korea and NWPU-RESISC45 datasets, and the results validate the effectiveness of attentive group convolution triplet loss network for retrieval tasks. Retrieval results are calculated for various state-of-the-art methods in the original and rotated datasets. R-MAC descriptor, NetVLAD, and LDCNN use distinct methods to extract critical features for RSIR task. R-MAC descriptor has sum-pooling to extract localization information from images. NetVLAD and LDCNN utilize mlpconv layers to focus on local descriptors. However, the key difference between these methods and attentive G-CNN is the rotated filters and attenion module used to extract channel refined rotation invariant features. The attentive group convolution triplet loss network with p4m group convolution layers has the best performance followed by the attentive group convolution triplet loss network with p4 group convolution layers. These results are not surprising because the attentive group convolution layers produce more feature maps and better training compared to regular convolution layers. P4m produces 8 channels of filters, p4 forms 4, and original convolution yields 1. This is why the network using p4m group convolution layers show superior results, followed by p4 group convolution and original convolution. 

\section{Conclusion}
We propose an attentive group convolutional metric learning network for rotation invariant remote sensing image retrieval. The advantage of this model is being able to retrieve remote sensing images with rotational variations between the query and database images. We design a two-step process using group convolution layers to train the network for classification and then fine-tuning for retrieval. This enables our model to be able to achieve better performance compared to other state-of-the-art methods in both original and rotated datasets. 

\section{Acknowledgment}
This work was supported by Institute of Information \& communications Technology Planning \& Evaluation (IITP) grant funded by the Korea government(MSIT) (No. 2019-0-00079, Artificial Intelligence Graduate School Program, Korea University).

\bibliographystyle{IEEEtran}
\bibliography{bib.bib}


\end{document}